\documentclass[10pt,twocolumn,letterpaper]{article}

\usepackage{cvpr}              
\usepackage{booktabs}
\usepackage{multirow}
\usepackage{graphicx}
\usepackage{algorithm}
\usepackage{algorithmic}

\definecolor{cvprblue}{rgb}{0.21,0.49,0.74}
\usepackage[pagebackref,breaklinks,colorlinks,allcolors=cvprblue]{hyperref}
\usepackage{hyperref}
\usepackage[hyphens]{url}          
\usepackage{multirow}

\hypersetup{colorlinks=true,linkcolor=red,citecolor=blue,urlcolor=magenta}
\newcommand{\smallurl}[1]{\footnotesize\url{#1}}

\usepackage{pifont}

\def\paperID{*****} 
\def\confName{CVPR}
\def\confYear{2026}

\title{Robust Fair Disease Diagnosis in CT Images}

\author{
Justin Li\textsuperscript{1}\thanks{Co-first Authors}\quad
Daniel Ding\textsuperscript{1}$^*$ \quad
Asmita Yuki Pritha \textsuperscript{2}$^*$\quad
Aryana Hou \textsuperscript{3}$^*$\quad
Xin Wang \textsuperscript{4}\quad 
Shu Hu\textsuperscript{5}\thanks{Corresponding author.}\\
\textsuperscript{1} Carmel High School, Carmel, Indiana, USA {\tt\small \{justinyli2018,danielding56\}@gmail.com}\\
\textsuperscript{2} Capstone School Dhaka, Bangladesh {\tt\small  asmitayukipritha@gmail.com}\\
\textsuperscript{3} Clarkstown High School South, West Nyack, New York, USA  {\tt\small  aryanahou@gmail.com}
\\
\textsuperscript{4} University at Albany, State University of New York, Albany, New York, USA  {\tt\small  xwang56@albany.edu}
\\
\textsuperscript{5}Purdue University, West Lafayette, Indiana, USA
{\tt\small hu968@purdue.edu}
}

\begin{document}
\maketitle

\begin{abstract}
Automated diagnosis from chest CT has improved considerably with deep learning, but models trained on skewed datasets tend to perform unevenly across patient demographics. However, the situation is worse than simple demographic bias. In clinical data, class imbalance and group underrepresentation often coincide, creating compound failure modes that neither standard rebalancing nor fairness corrections can fix alone. We introduce a two-level objective that targets both axes of this problem. Logit-adjusted cross-entropy loss operates at the sample level, shifting decision margins by class frequency with provable consistency guarantees. Conditional Value at Risk aggregation operates at the group level, directing optimization pressure toward whichever demographic group currently has the higher loss. We evaluate on the Fair Disease Diagnosis benchmark using a 3D ResNet-18 pretrained on Kinetics-400, classifying CT volumes into Adenocarcinoma, Squamous Cell Carcinoma, COVID-19, and Normal groups with patient sex annotations. The training set illustrates the compound problem concretely: squamous cell carcinoma has 84 samples total, 5 of them female. The combined loss reaches a gender-averaged macro F1 of 0.8403 with a fairness gap of 0.0239, a 13.3\% improvement in score and 78\% reduction in demographic disparity over the baseline. Ablations show that each component alone falls short. The code is publicly available at \url{https://github.com/Purdue-M2/Fair-Disease-Diagnosis}
\end{abstract}

\section{Introduction}
\begin{figure}[t]
\centering
\includegraphics[width=\columnwidth]{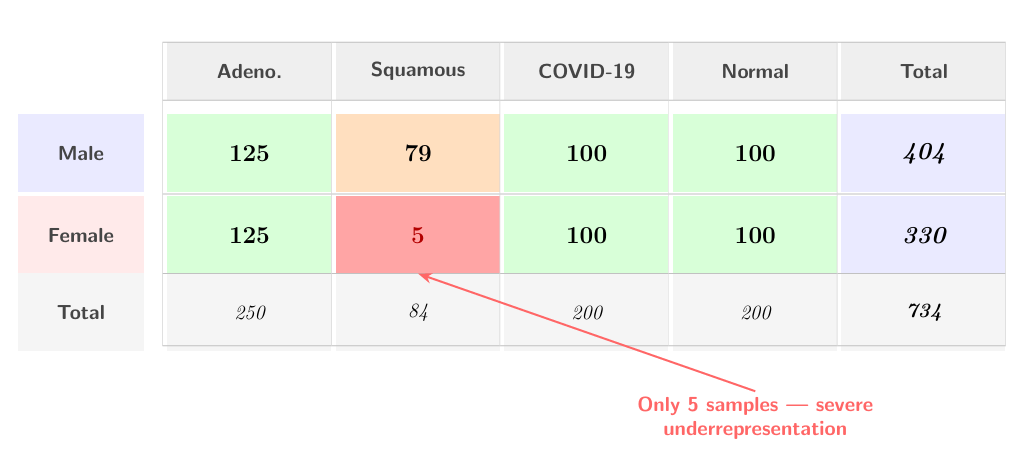}
\caption{\textbf{Training set distribution by disease category and patient sex.} The dataset exhibits a compound imbalance: squamous cell carcinoma is already the rarest class overall (84 samples), and its female subset contains only 5 training examples. This intersection of class rarity and demographic scarcity motivates our two-level objective that simultaneously addresses both axes of imbalance.}
\label{fig:distribution}
\end{figure}
Lung disease remains one of the leading causes of death worldwide, and the diagnostic landscape has only gotten more complicated in recent years. Lung cancer causes more than 1.8 million deaths annually. Among the non-small cell subtypes, adenocarcinoma and squamous cell carcinoma are the most common, and distinguishing between them matters because the treatment protocols diverge substantially~\cite{litjens2017survey}. In 2019, the COVID-19 began, quickly changing the lung disease field. The disease spreads rapidly, frequently present with no symptoms at all, yet produces detectable abnormalities on chest CT --- one study reported findings in 62.2\% of asymptomatic patients~\cite{inui2020chest}. The imaging infrastructure to screen for these conditions already exists and is widely implemented. Ninety-six percent of US emergency departments in a randomly sampled survey had a CT scanner on site~\cite{defined2017ctaccess}. The scanners are not the bottleneck; rather,  radiologists are. The volume of scans generated across institutions far exceeds what clinicians can read in a reasonable timeframe, and that throughput gap is what has driven the sustained push toward automated deep learning solutions~\cite{kollias2018deep, arsenos2022large}.
 
These solutions have matured quickly, and modern architectures on single-institution benchmarks achieve strong aggregate performance~\cite{kollias2023harmonizing, kollias2021miacov19d}. But aggregate performance is a dangerous metric when the underlying data has a structure that the model can exploit unevenly. Clinical CT data consistently has this kind of structure, and the specific problem we address is the interaction between class imbalance and demographic skew. Rare conditions are underrepresented in training data because they are rare. That much is well known. What is less obvious is that the imbalance is not the same across patient groups. In our benchmark, squamous cell carcinoma has 84 training samples out of 734 total. Split by sex, 79 are male and 5 are female. A classifier trained with standard cross-entropy on this distribution will learn squamous cell features almost entirely from male presentations, because that is where the gradient originates from. It will then fail on female patients presenting with the same disease, and because the class is already small, the overall accuracy will scarcely register the drop. The failure is real but it hides in the aggregate~\cite{cao2019ldam, menon2021logit, xu2024fairness}.
 
The fairness and class-rebalancing literature have proposed sensible solutions to their respective halves of this problem, but they were not designed to handle the intersection. Logit adjustment~\cite{menon2021logit} adds frequency-dependent offsets to the logits, enlarging the margin for rare classes. Menon et al. proved Fisher consistency for the balanced error rate $\tau = 1$, a guarantee that neither LDAM~\cite{cao2019ldam} nor equalization loss~\cite{tan2020equalization} provides. But logit adjustment does not see group labels --- every patient gets the same offset regardless of sex, so if male squamous cell samples dominate the gradient, the correction still flows predominantly through male data. CVaR based fairness aggregation~\cite{ju2024improving, rockafellar2000cvar} works from the other direction, upweighting whichever demographic group currently has higher loss. The trouble is that when per-sample loss is itself class-imbalanced, CVaR equalizes a biased signal. Both groups get similar loss values, but within each group the rare class is still being ignored. Neither mechanism, applied alone, can reach the particular intersection of the distribution where the real clinical risk sits.
 
We propose combining these methods into a single two-level objective. Logit-adjusted cross-entropy handles class imbalance at the sample level by reshaping which classes contribute to the gradient. CVaR aggregation handles demographic equity at the group level by reshaping which groups the optimizer prioritizes. The two mechanisms act on different dimensions of the loss function --- one controls per-sample gradient direction along the class axis, the other controls per-group gradient magnitude along the demographic axis --- and their composition is what allows the gradient to finally reach the specific intersection that neither can reach alone. We implement this on a 3D ResNet-18 pretrained on Kinetics-400 and evaluate on the Fair Disease Diagnosis benchmark of the PHAROS-AIF-MIH CVPR Workshop~\cite{kollias2024domain, kollias2025pharos}, a four class lung CT classification task with explicit gender equity requirements.
 
Our contributions are as follows:
\begin{enumerate}[leftmargin=*, label=(\arabic*), itemsep=2pt, topsep=0pt, parsep=0pt]
    \item We propose a two level training objective combining logit adjusted cross entropy with CVaR group aggregation, and show that the two components address orthogonal failure modes whose intersection produces the most clinically dangerous errors in the dataset.

    \item We experiment and analyze the effectiveness of our proposed approach and compare it against a baseline to evaluate the contributions of our method.
    
    
\end{enumerate}

\section{Related Work}
\subsection{Disease Diagnosis in CT Images}
 
Deep learning has become the prevailing approach for automated disease diagnosis from computed tomography, with applications spanning lung cancer subtype classification, COVID-19 screening, and pulmonary abnormality detection~\cite{litjens2017,kollias2018,yang2024explainable}. Early methods \cite{zhu2024cgd} decomposed 3D CT volumes into individual 2D slices and classified them using pretrained convolutional networks such as ResNet~\cite{he2016}, DenseNet~\cite{huang2017}, and Xception~\cite{chollet2017}. These approaches perform reasonably well when training and testing occur on data from the same institution. However, they discard spatial continuity between consecutive slices, which is critical for identifying diffuse pathologies such as ground glass opacities and for differentiating cancer subtypes that appear similar in any individual slice~\cite{arsenos2022,kollias2021}.
 
This limitation motivated the shift toward 3D convolutional architectures, typically pretrained on video data~\cite{hara2018,tran2018}. Kollias~et~al.~\cite{kollias2023} introduced RACNet, a routing and alignment architecture designed to handle CT volumes of varying length within a unified decision framework, and established a strong benchmark on COV19-CT-DB. Li~et~al.~\cite{li2025} later achieved competitive multi-class results on the same dataset using 3D ResNeSt50~\cite{zhang2022}. Neither method, however, models how data distributions vary across hospitals or acquisition sites; both formulate diagnosis as a single-task problem.
 
Multi-task learning (MTL) provides a principled remedy by optimizing several related objectives simultaneously, encouraging the shared backbone to learn features that generalize across tasks~\cite{caruana1997}. MTL has yielded consistent gains elsewhere in medical imaging, including joint segmentation and classification in chest radiography~\cite{chen2020} and simultaneous lesion detection and attribute prediction in dermoscopy~\cite{kawahara2019}, yet CT diagnostic pipelines have largely not adopted it. Pairing a disease classification head with an auxiliary source-identification head could force the network to learn site invariant representations, but this configuration has not been explored in a setting that must also contend with class imbalance and demographic fairness.
 
Two additional challenges compound these limitations. CT datasets are almost always imbalanced: common conditions dominate while rare but clinically significant pathologies such as squamous cell carcinoma are underrepresented~\cite{cao2019,menon2021}. At the same time, growing evidence indicates that diagnostic models perform unevenly across patient subgroups stratified by gender~\cite{xu2024,riccilara2022,kollias2024}. These issues are not independent, as different demographic groups tend to exhibit different disease prevalences, so mitigating one without accounting for the other risks exacerbating the remaining disparity. Current CT frameworks either overlook both problems or treat them in isolation.
 
\subsection{Fair Machine Learning and Class Imbalance}
 
There is now substantial evidence that models trained on demographically skewed data produce systematically disparate outcomes for subgroups defined by gender, race, or age~\cite{mehrabi2021,riccilara2022, ding2026decoupling,hou2025rethinking,bansal2025robust,wu2025preserving,lin2024preserving,hu2024fairness,ju2024improving,hu2022distributionally}. In medical imaging the implications are particularly serious: Xu~et~al.~\cite{xu2024} demonstrated that aggregate accuracy can conceal significant subgroup-level failures across modalities ranging from chest radiography to cardiac MRI. Standard mitigation strategies include preprocessing techniques that rebalance group representation~\cite{nadimpalli2022}, in processing objectives that embed fairness constraints directly into learning~\cite{hardt2016,dwork2012}, and post-processing calibration applied after training~\cite{wang2022}.
 
Among in processing formulations, Conditional Value at Risk (CVaR)~\cite{rockafellar2000,levy2020} has attracted significant attention as it provides a tractable upper bound on worst case group risk. It belongs to the broader family of rank based decomposable losses~\cite{hu2023,hu2022a}, and Hu and Chen~\cite{huchen2022} established its clinical viability by achieving demographic fairness in survival analysis without requiring group labels. Ju~et~al.~\cite{ju2024} extended this line of work with DAW-FDD, a hierarchical CVaR loss whose outer level equalizes performance across demographic groups while the inner level handles class imbalance within each group. DAW-FDD delivers strong intra domain fairness, but it depends on explicit demographic annotations and has only been validated on binary detection tasks. Lin~et~al.~\cite{lin2024} identified a more fundamental shortcoming: fairness gains from such methods collapse under domain shift, as the model memorizes source specific demographic patterns rather than learning invariant representations. Their remedy, combining feature disentanglement, a bi level fairness loss, and sharpness aware minimization~\cite{foret2020}, improved cross domain fairness but introduced considerable architectural overhead and has not been evaluated in a clinical imaging context. Hou~et~al.~\cite{hou2025} raised a separate concern, arguing that group level metrics are insufficiently granular, and proposed anchor learning with semantic agnostic individual fair learning to address per sample equity.
 
On the class imbalance side \cite{krubha2025robust,lin2024robust0,lin2024robust,lin2024robust1,hu2023rank,pu2022learning,guo2022robust,hu2022sum,hu2020learning}, logit adjustment~\cite{menon2021} offers a Fisher consistent correction by shifting logits according to label frequencies~\cite{cao2019}, yet its interaction with fairness aware objectives remains unstudied. To address this gap, we unify a CVaR based fairness objective and logit adjusted cross entropy inside a multi task learning framework, jointly targeting the coupled class imbalance, demographic fairness, and source heterogeneity problem in CT based disease diagnosis.
 
 \section{Method}
\begin{figure*}[t]
\centering
\includegraphics[width=\textwidth]{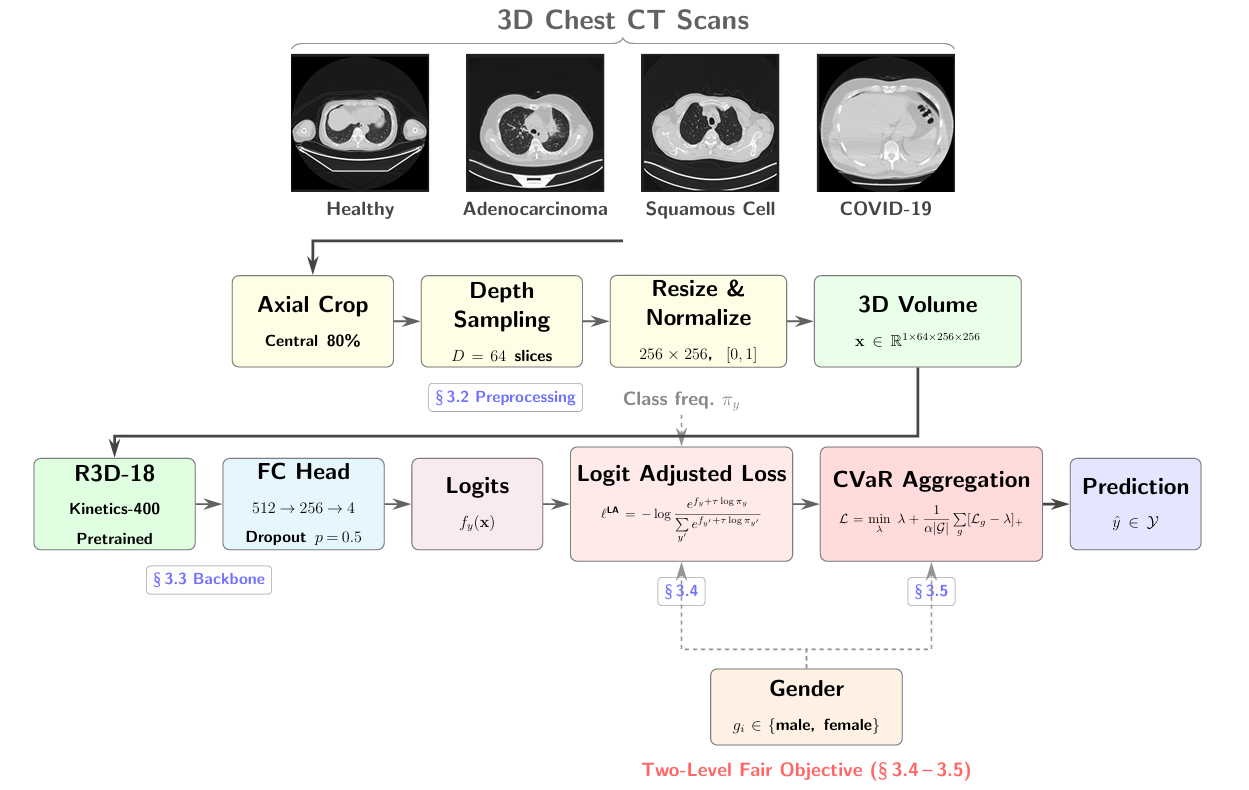}
\caption{\textbf{Overview of the proposed framework.} \textit{Row 1}: representative CT slices from the four diagnostic categories. \textit{Row 2}: preprocessing removes non-pulmonary regions, samples to a fixed depth of 64, and normalizes to produce the input volume. \textit{Row 3}: a Kinetics-400 pretrained R3D-18 extracts features, classified through a fully connected head. The logit adjusted loss (\S\,3.4) corrects class imbalance, while CVaR aggregation (\S\,3.5) enforces gender fairness. Dashed arrows indicate metadata inputs.}
\label{fig:framework}
\end{figure*}

\subsection{Problem Formulation}

We consider a training set $\mathcal{S} = \{(x_i, y_i, g_i)\}_{i=1}^{n}$, where each $x_i$ is a 3D chest CT volume, $y_i \in \mathcal{Y}=\{\text{Healthy}, \text{Adenocarcinoma}, \text{Squamous Cell}, \text{COVID-19}\}$ is the diagnostic label,
and $g_i \in \mathcal{G} = \{\text{male},\; \text{female}\}$ records patient sex. We write $\mathcal{I}_g = \{i \in [n] : g_i = g\}$ for the index set of group $g$.

We want a classifier $f_\theta\colon x \mapsto \mathcal{Y}$ that achieves two things at once: (i) strong diagnostic accuracy across all four categories despite severe class imbalance, and (ii) equitable performance across genders, measured by
\begin{equation}
P' = \frac{1}{2}\big(F1_{\text{male}}^{\text{macro}} + F1_{\text{female}}^{\text{macro}}\big).
\end{equation}

These objectives interact in ways that make tackling them separately insufficient. Squamous cell carcinoma, for instance, contributes as few as 5 female training samples. Na\"ive empirical risk minimization on such a distribution will push the decision boundaries toward dominant classes. But a group level fairness mechanism applied on top of an imbalanced loss just propagates the same class bias into both groups equally, which helps neither.

Our framework (Figure~\ref{fig:framework}) handles both problems through a two level objective: logit adjusted cross entropy~\cite{menon2021logit} at the sample level corrects class imbalance, while CVaR aggregation~\cite{ju2024improving} at the group level enforces demographic equity.

\subsection{Preprocessing}
CT scans in the dataset range from roughly 50 to over 700 axial slices at $512 \times 512$ resolution. We discard the top and bottom 10\% of slices to remove non pulmonary regions (cervical spine, upper abdomen), then uniformly sample 64 slices from the remainder via linear index interpolation. Each slice is converted to grayscale, resized to $256 \times 256$, and normalized to $[0, 1]$, producing an input tensor $x \in \mathbb{R}^{1 \times 64 \times 256 \times 256}$. Training augmentation consists of random $224 \times 224$ crops and horizontal flips; inference uses a center crop.

\subsection{Backbone Architecture}

Our backbone is R3D-18~\cite{tran2018closer} pretrained on Kinetics-400. Video pretrained spatiotemporal kernels transfer well to CT volumes because both domains exhibit spatially coherent patterns that evolve gradually along the third axis. We replace the first convolutional layer with a single channel $3 \times 7 \times 7$ kernel initialized by averaging the pretrained RGB weights along the channel dimension, and swap the classification head for a fully connected network ($512 \to 256 \to 4$) with ReLU and 0.5 dropout.

\subsection{Logit Adjusted Cross Entropy for Class Imbalance}

Standard cross entropy on a skewed distribution biases the classifier toward frequent classes. Inverse frequency reweighting is a common fix, but Byrd and Lipton~\cite{byrd2019effect} show it has limited effect in separable regimes because it does not change the inter class margins.

Logit adjustment~\cite{menon2021logit} works differently. Instead of reweighting the loss, it adds class frequency dependent offsets directly to the logits. Let $f_y(x)$ be the logit for class $y$ and $\pi_y$ its empirical frequency. The logit adjusted loss is
\begin{equation}
\ell^{\text{LA}}(x, y) = -\log \frac{\exp\big(f_y(x) + \tau \cdot \log \pi_y\big)}{\sum_{y' \in \mathcal{Y}} \exp\big(f_{y'}(x) + \tau \cdot \log \pi_{y'}\big)},
\label{eq:la}
\end{equation}
where $\tau > 0$ controls the adjustment strength. This is equivalently a pairwise margin loss with margin $\Delta_{yy'} = \tau \cdot \log(\pi_{y'}/\pi_y)$: when $y$ is rare and $y'$ is dominant, the margin grows large, forcing wide logit separation. This is exactly what inverse frequency weighting fails to achieve.

Menon et al.~\cite{menon2021logit} prove that with $\tau = 1$, this loss is Fisher consistent for the balanced error rate, meaning its minimizer coincides with the Bayes optimal scorer. Neither LDAM~\cite{cao2019learning} nor the equalization loss~\cite{tan2020equalization} provides this guarantee. We fix $\tau = 1$ throughout.

The probabilistic view is also useful: training with logit adjustment encourages estimation of the balanced posterior $P^{\text{bal}}(y \mid x) \propto P(y \mid x)/P(y)$ rather than the native $P(y \mid x)$, yielding predictions invariant to training set imbalance.

\subsection{CVaR Fairness Aggregation}

Logit adjustment corrects class imbalance but is completely blind to demographics. It applies the same offset to every patient regardless of sex, so if the model can reduce loss more easily on one gender, it will, and the per gender F1 gap persists.

We handle this with CVaR~\cite{rockafellar2000optimization, ju2024improving}, which aggregates group losses in a way that places more weight on whichever group is currently doing worse. For each $g \in \mathcal{G}$, the mean logit adjusted loss is
\begin{equation}
\mathcal{L}_g = \frac{1}{|\mathcal{I}_g|} \sum_{i \in \mathcal{I}_g} \ell_i^{\text{LA}}.
\label{eq:grouploss}
\end{equation}

The training objective is the group level CVaR:
\begin{equation}
\mathcal{L} = \min_{\lambda \in \mathbb{R}} \;\; \lambda + \frac{1}{\alpha\,|\mathcal{G}|} \sum_{g \in \mathcal{G}} [\mathcal{L}_g - \lambda]_+,
\label{eq:cvar}
\end{equation}
where $[\cdot]_+ = \max(0, \cdot)$ and $\alpha \in (0, 1]$ controls fairness strength. At $\alpha \to 0$ this becomes pure minimax over groups~\cite{rawls2001justice}; at $\alpha = 1$ it is a simple average. The optimal $\lambda$ is convex and found via binary search at each step with negligible overhead.

\subsection{Analysis of the Joint Objective}

The full procedure is given in Algorithm~\ref{alg:training}. The binary search at line 8 requires no gradient computation. The rest of the pipeline is end to end differentiable.
\begin{algorithm}[t]
\caption{Training with LA + CVaR}
\label{alg:training}
\begin{algorithmic}[1]
\REQUIRE Training set $\mathcal{S}$, group set $\mathcal{G}$, class frequencies $\{\pi_y\}$, $\alpha$, learning rate $\eta$
\STATE Initialize $\theta$ from Kinetics-400 pretrained R3D-18
\FOR{each epoch}
\FOR{each minibatch $\mathcal{B} \subset \mathcal{S}$}
\STATE Compute $\ell_i^{\text{LA}}$ for all $i \in \mathcal{B}$ \hfill $\triangleright$ Eq.~(\ref{eq:la})
\FOR{each $g \in \mathcal{G}$}
\STATE $\mathcal{L}_g \gets \frac{1}{|\mathcal{I}_g \cap \mathcal{B}|} \sum_{i \in \mathcal{I}_g \cap \mathcal{B}} \ell_i^{\text{LA}}$ \hfill $\triangleright$ Eq.~(\ref{eq:grouploss})
\ENDFOR
\STATE Find $\lambda^*$ via binary search on $\{\mathcal{L}_g\}_{g \in \mathcal{G}}$ \hfill $\triangleright$ convex in $\lambda$
\STATE $\mathcal{L} \gets \lambda^* + \frac{1}{\alpha|\mathcal{G}|} \sum_{g \in \mathcal{G}} [\mathcal{L}_g - \lambda^*]_+$ \hfill $\triangleright$ Eq.~(\ref{eq:cvar})
\STATE $\theta \gets \theta - \eta \nabla_\theta \mathcal{L}$
\ENDFOR
\ENDFOR
\end{algorithmic}
\end{algorithm}

Is combining logit adjustment with CVaR just stacking two known components, or does the interaction produce something neither achieves alone? We argue the latter.

\textbf{Orthogonality.} Logit adjustment reshapes which classes receive gradient signal at the per sample level. It does not see group labels. CVaR reshapes which groups receive optimization pressure at the aggregate level. It does not see per class structure. CVaR alone on an imbalanced dataset may equalize group losses while both groups still ignore the rare class. The two mechanisms act on orthogonal axes: one controls per sample gradient direction (class axis), the other controls per group gradient magnitude (demographic axis).

\textbf{Complementary failure modes.} Squamous cell carcinoma has 79 male but only 5 female training samples. Logit adjustment alone upweights this class uniformly, but 94\% of the squamous cell gradient comes from male samples, so the learned features specialize to male presentation. CVaR alone detects the gender gap and shifts pressure toward the female group, but within that group the abundant classes dominate and the extra gradient never reaches the rare class. Only the composition routes sufficient signal to the precise intersection (female, squamous cell carcinoma) that is most at risk.

\textit{\textbf{Remark 1.} Logit adjustment is invariant to group membership; CVaR is invariant to per class structure. Their composition is the minimal objective that is simultaneously sensitive to both the class axis and the demographic axis of the training distribution.}

\section{Experimental Settings}

\subsection{Datasets}

We evaluate on the Fair Disease Diagnosis benchmark~\cite{kollias2024domain, kollias2025pharos}, which contains 3D chest CT scans labeled across four categories: Adenocarcinoma (A), Squamous Cell Carcinoma (G), COVID-19, and Normal. Patient sex (male/female) is provided for each scan. The training set has 734 scans (404 male, 330 female) and the validation set has 155 (77 male, 78 female). Scan lengths range from roughly 50 to 700 slices at 512$\times$512 resolution. As shown in Figure~\ref{fig:distribution}, the class distribution is heavily skewed: squamous cell carcinoma has only 84 training samples in total, and just 5 of those are female. This severe overlap between class rarity and gender underrepresentation is what drives our fairness aware loss design.

\subsubsection{Evaluation Metrics}

We adopt the official protocol from~\cite{kollias2025pharos}. The validation set is split by sex into Subset A (male) and Subset B (female), and a four class macro F1 is computed on each. The final score averages the two:
\begin{equation}
P = \frac{1}{2}\left(\mathrm{F1}_{\mathrm{male}}^{\mathrm{macro}} + \mathrm{F1}_{\mathrm{female}}^{\mathrm{macro}}\right)
\end{equation}
The macro F1 here is simply the unweighted mean of per class F1 scores. A model that performs well on males but poorly on females (or vice versa) will be penalized directly. We also report the fairness gap, $|\mathrm{F1}_{\mathrm{male}}^{\mathrm{macro}} - \mathrm{F1}_{\mathrm{female}}^{\mathrm{macro}}|$, as a scalar measure of demographic disparity.

\subsubsection{Baseline Methods}

We compare FairLACVaR against the competition baseline and three ablation variants that isolate each component of our loss. \textbf{(1) Competition Baseline (CNN-RNN):} The organizer baseline~\cite{kollias2025pharos} pairs a ResNet-50 encoder with a unidirectional GRU (128 units) for slice level aggregation. It is trained with softmax cross-entropy using gender balanced loss averaging, processes scans padded to 700 slices, and achieves a score of 0.6230. \textbf{(2) Baseline (CE):} Our own 3D ResNet-18 trained with standard cross-entropy and no fairness mechanism at all. This is the direct ablation target. \textbf{(3) LA Only:} Same architecture, but trained with Logit-Adjusted cross-entropy~\cite{menon2021logitadjust} to account for class frequency skew. No CVaR aggregation across genders. \textbf{(4) CVaR Only:} Standard cross-entropy wrapped in CVaR based group fairness aggregation ($\alpha = 0.7$) over the two gender groups, without logit adjustment. These ablations cleanly separate what class aware calibration contributes from what demographic aware optimization adds. As Table~\ref{tab:main} shows, neither LA nor CVaR alone reaches the score or the fairness gap of the combined FairLACVaR loss, which confirms that both problems need to be tackled together.

\subsubsection{Implementation Details}

All experiments run in PyTorch on a single NVIDIA A100 GPU. The backbone is a 3D ResNet-18~\cite{he2016deep} pretrained on Kinetics-400~\cite{kay2017kinetics}. Since CT scans are grayscale, we collapse the first convolution from 3 input channels to 1 by averaging the pretrained RGB weights along the channel axis. On top of the encoder we attach a classification head: a linear layer (512$\rightarrow$256), ReLU, dropout ($p = 0.5$), and a final four class projection.

For preprocessing, we strip the top and bottom 10\% of slices from each scan to get rid of neck and abdominal regions that carry no diagnostic signal. The remaining volume is resampled to 64 slices and resized to 256$\times$256, with intensities normalized to $[0, 1]$. Training augmentations include random cropping to 224$\times$224, random 90$^{\circ}$ rotations in the axial plane, and brightness jittering (scale factor sampled from $[0.8, 1.2]$). At test time we use a fixed center crop of 224$\times$224.

We train for 70 epochs using Adam~\cite{kingma2015adam} with a learning rate of $1 \times 10^{-4}$ and weight decay of $1 \times 10^{-5}$, decayed via cosine annealing ($T_{\mathrm{max}} = 70$). Mixed precision is enabled through PyTorch's GradScaler. Batch size is 2, since full 3D volumes consume substantial GPU memory. We save every checkpoint and select the best by validation macro F1. All runs are seeded at 42 for reproducibility. In the FairLACVaR loss, we fix the logit adjustment temperature at $\tau = 1.0$ and sweep the CVaR concentration $\alpha$ over $\{0.4, 0.5, 0.6, 0.7, 0.8, 0.9\}$; the resulting trade-off between overall score and fairness gap is reported in Table~\ref{tab:main}.

\subsection{Results}

\begin{figure}[t]
\centering
\includegraphics[width=\columnwidth]{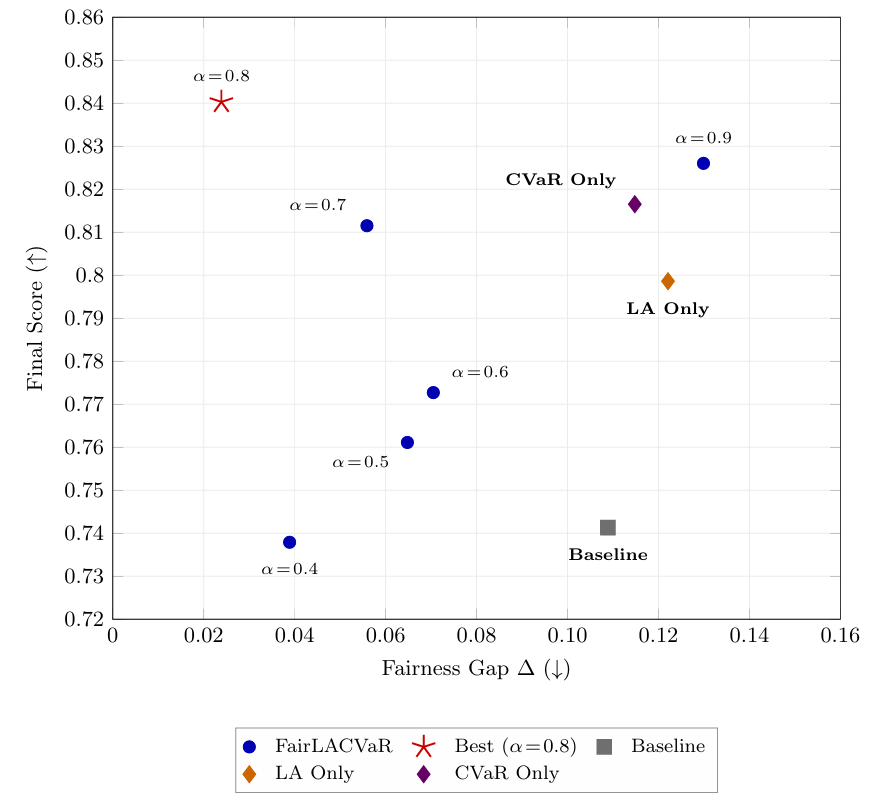}
\caption{Performance vs.\ fairness trade-off across all configurations. Circles ($\bullet$) denote FairLACVaR at different $\alpha$ values, diamonds denote single-component ablations, and the square marks the baseline. The star ($\star$) indicates the best overall configuration ($\alpha = 0.8$), which occupies the upper-left region corresponding to high score and low fairness gap.}
\label{fig:tradeoff}
\end{figure}

\begin{table}[t]
\centering
\caption{Main results and ablation study on the challenge validation set. All runs use a fixed random seed on NVIDIA A100. Best values in \textbf{bold}.}
\label{tab:main}
\setlength{\tabcolsep}{3pt}
\resizebox{\columnwidth}{!}{%
\begin{tabular}{@{}lccccc@{}}
\toprule
Method & $\alpha$ & F1\textsubscript{male} & F1\textsubscript{female} & Score $\uparrow$ & Gap $\downarrow$ \\
\midrule
\multicolumn{6}{@{}l}{\textit{Baseline \& Ablations}} \\
Baseline (CE) & -- & 0.7957 & 0.6868 & 0.7413 & 0.1089 \\
LA Only & -- & 0.8596 & 0.7375 & 0.7986 & 0.1221 \\
CVaR Only & 0.7 & 0.8738 & 0.7591 & 0.8165 & 0.1148 \\
\midrule
\multicolumn{6}{@{}l}{\textit{FairLACVaR (LA + CVaR)}} \\
LA + CVaR & 0.4 & 0.7574 & 0.7185 & 0.7379 & 0.0389 \\
LA + CVaR & 0.5 & 0.7935 & 0.7287 & 0.7611 & 0.0648 \\
LA + CVaR & 0.6 & 0.8079 & 0.7374 & 0.7727 & 0.0705 \\
LA + CVaR & 0.7 & 0.7835 & 0.8394 & 0.8115 & 0.0559 \\
LA + CVaR & 0.8 & 0.8283 & \textbf{0.8522} & \textbf{0.8403} & \textbf{0.0239} \\
LA + CVaR & 0.9 & \textbf{0.8910} & 0.7611 & 0.8260 & 0.1299 \\
\bottomrule
\end{tabular}%
}
\end{table}

We sweep the CVaR confidence level $\alpha$ from 0.4 to 0.9 on the challenge validation set under a fixed random seed, with all runs on a single NVIDIA A100. Table~\ref{tab:main} reports results alongside the baseline and ablation variants.

The baseline, trained with standard cross-entropy, scores 0.7413 with a gap of 0.1089. When we look at this by class and gender, the problem is clear: female Squamous Cell recall is just 0.08 against 0.75 for males. With only 5 female Squamous Cell training samples versus 79 male, the network maps most of these cases onto Adenocarcinoma instead, which shows up as low female Adenocarcinoma precision (0.68).

With $\alpha = 0.8$, our combined loss reaches 0.8403 with a gap of 0.0239, a 13.3\% score improvement and 78\% gap reduction over the baseline. Female Squamous Cell F1 goes from 0.14 to 0.63 as recall rises from 0.08 to 0.46 at a precision of 1.00. COVID-19 and Normal F1 stay above 0.90 for both genders across all settings. This is also the only configuration where female macro F1 (0.8522) exceeds male (0.8283), so the model closes the gap by lifting the underperforming group rather than dragging down the majority.

Two ablations (Table~\ref{tab:main}) confirm that both components are needed. Logit adjustment alone improves score (0.7986) but widens the gap (0.1221). CVaR alone reaches 0.8165 but leaves the gap at 0.1148. Neither one handles the interaction between class and demographic imbalance. The combined loss at $\alpha = 0.8$ beats LA-only by 5.2\% in score with 80\% gap reduction, and CVaR-only by 2.9\% with 79\% gap reduction. Logit adjustment corrects class-frequency bias, CVaR shifts gradient emphasis to the hardest subgroup samples, and removing either leaves a gap in coverage.

The full sweep shows a non-monotonic pattern across three phases (Figure~\ref{fig:tradeoff}). From $\alpha = 0.4$ to 0.6, score rises (0.7379 to 0.7727) but female F1 stays near 0.73 because the broad CVaR tail dilutes the fairness signal across classes. At $\alpha = 0.7$ the tail narrows enough to be dominated by hard female Squamous Cell cases and female F1 jumps to 0.8394, exceeding male F1 for the first time. This peaks at $\alpha = 0.8$. At $\alpha = 0.9$ the tail becomes too thin (worst 10\%), male F1 spikes to 0.8910, female F1 drops to 0.7611, and the gap reverts to 0.1299.

\subsection{Sensitivity Analysis}

\begin{table}[t]
\centering
\caption{Hardware sensitivity analysis. Same seed and code on NVIDIA A100 vs.\ L4. The optimal $\alpha$ shifts across platforms, but all configurations within $\alpha \in [0.7, 0.9]$ improve over the baseline.}
\label{tab:hardware}
\setlength{\tabcolsep}{3pt}
\resizebox{\columnwidth}{!}{%
\begin{tabular}{@{}clccccc@{}}
\toprule
$\alpha$ & GPU & F1\textsubscript{male} & F1\textsubscript{female} & Score $\uparrow$ & Gap $\downarrow$ \\
\midrule
\multirow{2}{*}{0.7}
 & A100 & 0.7835 & 0.8394 & 0.8115 & 0.0559 \\
 & L4 & 0.8467 & 0.7415 & 0.7941 & 0.1052 \\
\midrule
\multirow{2}{*}{0.8}
 & A100 & 0.8283 & \textbf{0.8522} & \textbf{0.8403} & \textbf{0.0239} \\
 & L4 & 0.7498 & 0.7591 & 0.7544 & 0.0093 \\
\midrule
\multirow{2}{*}{0.9}
 & A100 & 0.8910 & 0.7611 & 0.8260 & 0.1299 \\
 & L4 & 0.8379 & 0.7813 & 0.8096 & 0.0565 \\
\bottomrule
\end{tabular}%
}
\end{table}

We rerun $\alpha \in \{0.7, 0.8, 0.9\}$ on an NVIDIA L4 with the same seed and code (Table~\ref{tab:hardware}). We focus on the high-$\alpha$ range because the broader tail at lower $\alpha$ smooths the loss surface and should be less affected by hardware differences.

The results change considerably. On the A100, $\alpha = 0.8$ is the clear best (score 0.8403, gap 0.0239, epoch 69). On the L4, the same setting stops early at epoch 44 and reaches only 0.7544, with its near-zero gap (0.0093) coming from male F1 dropping rather than female F1 rising. At $\alpha = 0.9$ it flips: the L4 gives a balanced result (0.8096, gap 0.0565) while the A100 produces the worst gap in the study (0.1299). The optimal $\alpha$ shifts from 0.8 to 0.9 depending on the GPU.

This happens because the A100 and L4 use different cuDNN convolution kernels and accumulate rounding errors differently. These differences are tiny on their own but compound through backpropagation over training. At high $\alpha$ the CVaR surface is sharp with several competing minima that differ in fairness properties, so small numerical perturbations can push the model toward a fundamentally different solution.

Two things hold regardless of platform: the combined loss beats the baseline and both ablations in every tested configuration, and all settings within $\alpha \in [0.7, 0.9]$ achieve scores above 0.79 with gaps below 0.11. We recommend tuning $\alpha$ within this range on the target deployment hardware.

\section{Conclusion}
We presented a two level objective for fair disease diagnosis in chest CT that combines logit adjusted cross entropy at the sample level with CVaR aggregation at the group level. On the Fair Disease Diagnosis benchmark the combined loss achieves a gender averaged macro F1 of 0.8403 and a demographic gap of 0.0239, improving the baseline by 13.3\% in score and reducing the fairness disparity by 78\%. Neither component achieves this result independently: logit adjustment improves accuracy but widens the gender gap because 94\% of the rare class gradient flows through male data, while CVaR narrows the group disparity but cannot fix what it does not see --- the rare class stays neglected in both groups.
 
There are limitations to acknowledge, some more practical than others. The fairness formulation considers binary sex groups only. Real clinical populations vary along many dimensions --- age, ethnicity, comorbidity burden --- and extending CVaR to intersections of several attributes is not straightforward because the group structure grows combinatorially and the data requirements per cell grow with it. A more immediate issue is hardware sensitivity. We found that the optimal $\alpha$ shifts between GPUs because different hardware accumulates rounding errors differently through backpropagation, and at high $\alpha$ the CVaR surface has competing minima that differ meaningfully in fairness properties. The practical recommendation is to tune $\alpha$ on the deployment hardware rather than transferring a value from a different setup.
 
Future work will explore hierarchical CVaR structures for multiple sensitive attributes and evaluate whether the framework generalizes under cross institutional domain shift.

\section*{Acknowledgements}
This work is supported by the U.S. National Science Foundation (NSF) under grant IIS-2434967, and the National Artificial Intelligence Research Resource (NAIRR) Pilot and TACC Lonestar6. The views, opinions and/or findings expressed are those of the author and should not be interpreted as representing the official views or policies of NSF and NAIRR Pilot.

{
    \small
    \bibliographystyle{IEEEtran}
    \bibliography{main}
}

\end{document}